\documentclass[letterpaper, 10 pt, conference]{format/ieeeconf}
\usepackage{amsmath}
\usepackage{amsfonts}
\usepackage{setspace}
\usepackage[space, compress, sort]{cite}

\usepackage{amsthm}
\usepackage{amssymb,stmaryrd}
\usepackage[ruled,vlined]{algorithm2e}
\usepackage{mathtools}
\usepackage{tabularx}
\usepackage{graphicx}
\usepackage{subcaption}
\usepackage{enumerate}
\usepackage{float}
\usepackage{url}
\usepackage{verbatim}
\usepackage{booktabs} 
\usepackage{multirow}
\usepackage[linkcolor=black,citecolor=black,urlcolor=black,colorlinks=true]{hyperref}
\usepackage{bm}
\usepackage{empheq}
\IEEEoverridecommandlockouts
\usepackage{graphicx}
\usepackage{siunitx}
\usepackage{textcase}
\usepackage{dblfloatfix}
\usepackage[tablename=TABLE]{caption}
\DeclareCaptionTextFormat{up}{\MakeTextUppercase{#1}}

\title{\LARGE \bf
Developing Path Planning with Behavioral Cloning and Proximal Policy Optimization for Path-Tracking and Static Obstacle Nudging}

\author{Mingyan Zhou$^{1}$, Biao Wang$^{1}$, Tian Tan$^{1}$, Xiatao Sun$^{2}$ 
\thanks{$^{1}$ Mingyan Zhou and Tian Tan are with the Department of Electrical and Systems Engineering, Biao Wang is with the Department of Mechanical Engineering and Applied Mechanics, University of Pennsylvania, Philadelphia, PA 19104, USA (e-mail: \{derekzmy, wangbiao\}@alumni.upenn.edu, tiantan@seas.upenn.edu).}
\thanks{$^{2}$ Xiatao Sun is with the Department of Computer Science, Yale University, New Haven, CT 06510, USA (e-mail: xiatao.sun@yale.edu).}
}

\begin{document}

\maketitle
\thispagestyle{empty}
\pagestyle{empty}


\begin{abstract}
In autonomous driving, end-to-end methods utilizing Imitation Learning (IL) and Reinforcement Learning (RL) are becoming more and more common. 
However, they do not involve explicit reasoning like classic robotics workflow and planning with horizons, resulting in strategies implicit and myopic. 
In this paper, we introduce a path planning method that uses 
Behavioral Cloning (BC) for path-tracking and Proximal Policy Optimization (PPO) for static obstacle nudging. 
It outputs lateral offset values to adjust the given reference waypoints and performs modified path for different controllers. 
Experimental results show that the algorithm can do path following that mimics the expert performance of path-tracking controllers, and avoid collision to fixed obstacles. The method makes a good attempt at planning with learning-based methods in path planning problems of autonomous driving. 
\end{abstract}

\section{INTRODUCTION}

Reinforcement Learning (RL) and Imitation Learning (IL) has become more and more popular in the field of robotics. 
RL is a machine learning paradigm in which an agent progressively learns optimal decision-making strategies through repeated interactions with its environment. 
It has been improved significantly in recent years, from Vanilla Policy Gradient (VPG) \cite{sutton1999policy} to most widely-used RL method, Proximal Policy Optimization (PPO) \cite{schulman2017ppo}. 
Imitation Learning (IL) involves training an agent to replicate behaviors from expert demonstrations \cite{ahmed2017survey} rather than learning through direct interaction with the environment. 
Behavioral Cloning (BC), a foundational IL method, was initially introduced by \cite{sammut2010bc} using a supervised learning framework. 

Beyond theoretical developments, numerous robotics applications have demonstrated the effectiveness of both RL and IL.
These applications include quadrotors \cite{sun2022}, autonomous vehicles \cite{shenyi2024}, and robotic arms \cite{zhang2024self}. 
Several robotics platforms have been specifically developed for research purposes in autonomous driving, utilizing RL and IL methods, including AutoRally \cite{Pan2018DIL} for off-road driving and \cite{Cai2021DIRL} for validating RL-based algorithms. 
Among the various autonomous driving platforms, F1TENTH \cite{okelly2020f1tenth} stands out as one of the most promising. It features a reproducible simulation environment for rapid implementation and a wealth of open-source materials accumulated from extensive developer contributions. 
Significant research have been achieved using F1TENTH, including high-speed control \cite{10161472}, generalized RL \cite{bosello2022train}, and safe overtaking \cite{megadagger}.


In autonomous driving, classic software pipelines are modularly developed, components such as perception, planning and control \cite{betz2022autonomous} \cite{av4ev}.
Besides robotics pipelines, end-to-end methods are gaining prominence in autonomous driving research due to their simplified, efficient design and potential for continuous optimization.
End-to-end approaches replace either some or all of the software modules with data-driven methods, including various RL and IL techniques \cite{le2022survey}. 
For end-to-end approaches developed on the F1TENTH, \cite{zzj_f110rl} demonstrate a successful implementation of PPO, using downsampled lidar data as input and producing steering and speed commands as output. 
Similarly, \cite{f110IL} illustrates the feasibility and the benchmark comparison of Direct Policy Learning methods. 
However, two challenges remain to be addressed. 
First, the explainability of these methods is limited, as deep learning models often function as black boxes, making it difficult to interpret their decision-making processes. 
This lack of transparency complicates reasoning and further validation. 
Second, these approaches tend to exhibit myopic behavior, as they fail to adequately account for the planning horizon, causing the vehicle to reactively respond to inputs. 
As observed in \cite{follow_the_gap}, this can lead to problematic situations, especially on right-angle tracks and in real-world environments, despite efforts to mitigate these issues through various adjustments. 

To address the implicit and myopic issue, we propose the "planning with learning" method in this study. 
For opaque decision-making, we replaced the planning module within the autonomous driving software pipeline similarly in \cite{weiss2020deepracing} and \cite{lipeng2024prioritized}, narrowing down the learning-based approach to particular tasks. 
To enhance path planning with RL and IL to move beyond reactive performance, we shifted outputs from steering and speed commands to sequences of data such as lateral offsets or a series of actions. This enables a prediction for motion planning in a longer term. 
Additionally, based on the prior research in \cite{f110IL}, we utilized the strong bootstrapping capabilities of IL methods for achieving fast convergence and improved performance in the static obstacle nudging.
We showed that this method can mimic the expert demonstration of path-tracking controllers, and enables the controllers with static obstacle nudging features.  

\begin{figure}[!t]
\includegraphics[width=\columnwidth]{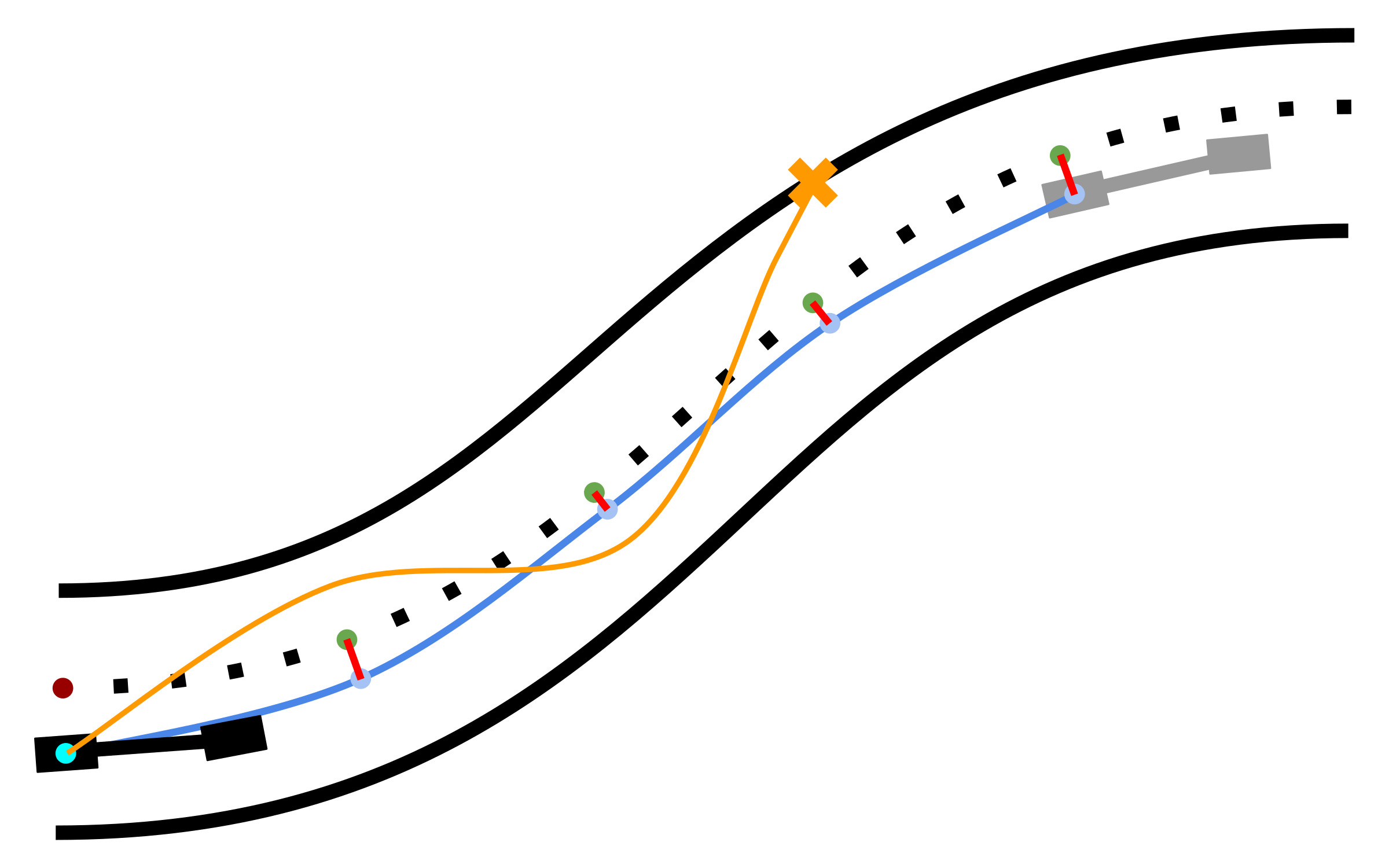} 
\caption{Tracking path through path-planning-based BC. Given demonstration (blue trace) from the expert (single-track model in grey), instead of deviation or collision (yellow marks), the vehicle (single-track model in black) learns to mimic the expert by adjusting lateral offsets (red line segments) on the selected path (green dots) obtained by current state (cyan dot), reference waypoints (black dots), and closest waypoint (crimson dot). }
\label{fig:illu_tracking}
\end{figure}
\begin{figure}[!t]
\includegraphics[width=\columnwidth]{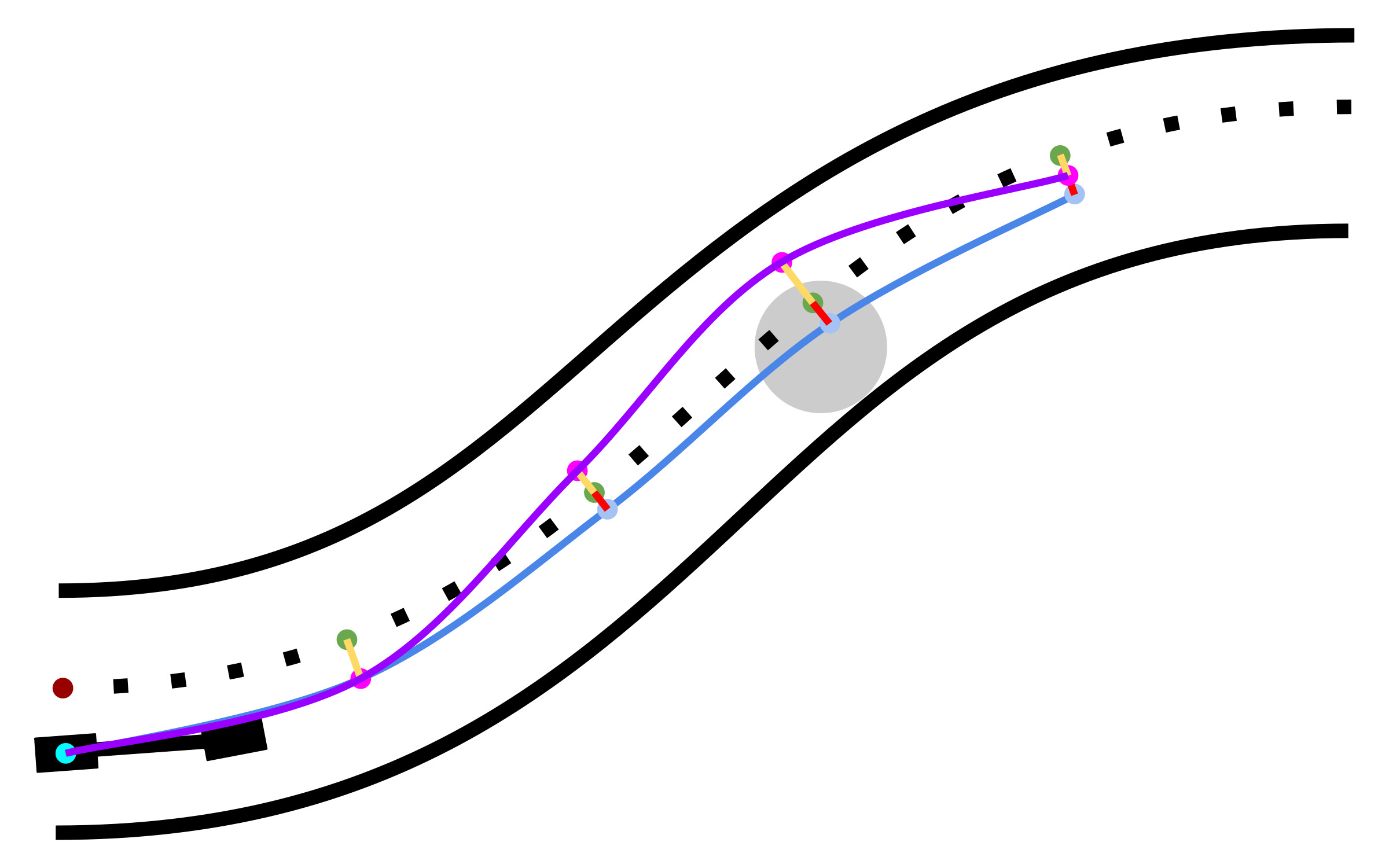} 
\caption{Static obstacle nudging through path-planning-based PPO. After bootstrapping by BC as illustrated in Fig. \ref{fig:illu_tracking}, the vehicle performs planning similar to the expert's (blue trace). To avoid obstacles (grey circle) that may block the path, the vehicle adopts PPO to adjust the policy that outputs offsets to get a new path (purple trace), which reflects as adding new deviations (yellow line segments) to get new waypoints (pink dots). }
\label{fig:illu_obs}
\end{figure}

This work makes three main contributions:
\begin{enumerate}
    \item We proposed a novel approach that improves planning to "planning with learning" as an integrated component of the robotics workflow for autonomous driving;
    \item We implemented the Behavioral Cloning (BC) algorithm on path planning for path-tracking, ensuring compatibility with various path-tracking controllers (Fig. \ref{fig:illu_tracking});
    \item We utilized the Proximal Policy Optimization (PPO) algorithm bootstrapped by BC to adjust reference waypoints with lateral offsets, enabling obstacle avoidance through nudging (Fig. \ref{fig:illu_obs}).
\end{enumerate}

\section{METHODOLOGY}

\begin{figure}[t]
\centering
\includegraphics[width=0.45\textwidth]{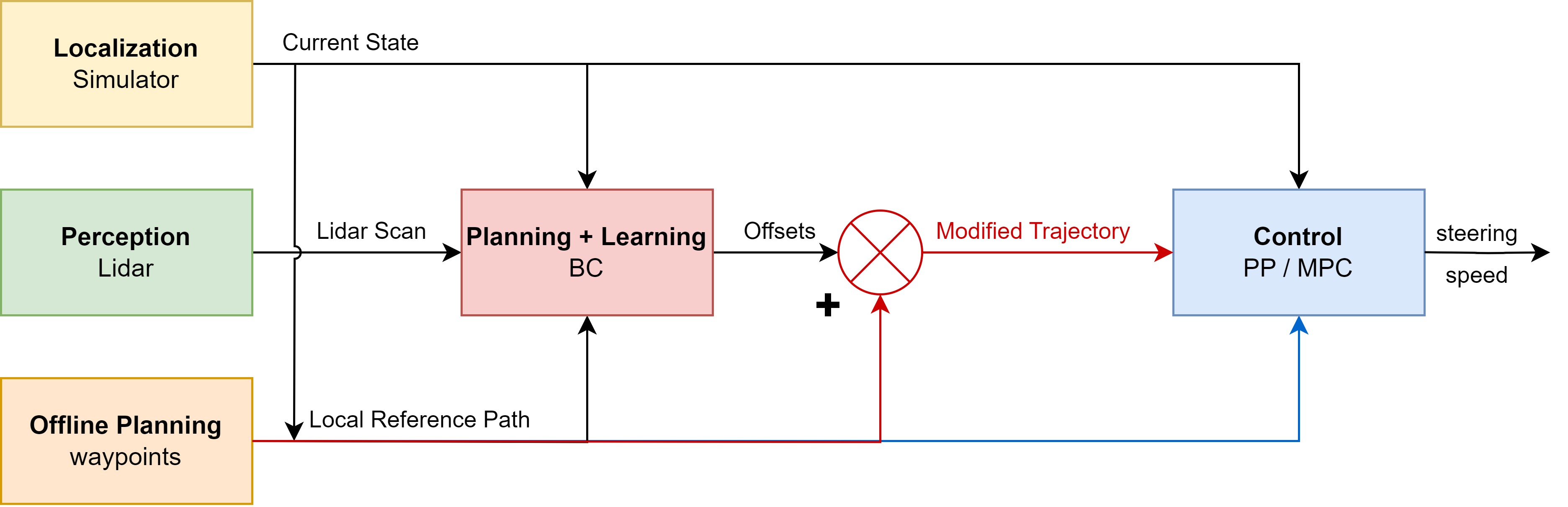} 
\caption{Structure of path-tracking with path-planning-based BC. Controller directly takes the path from waypoints as reference (blue lines) to train the policy. During validation process, offsets are added up the to get the modified path for controller (red lines). }
\label{fig:logic_IL_plot}
\end{figure}

\begin{figure}[t]
\centering
\includegraphics[width=0.45\textwidth]{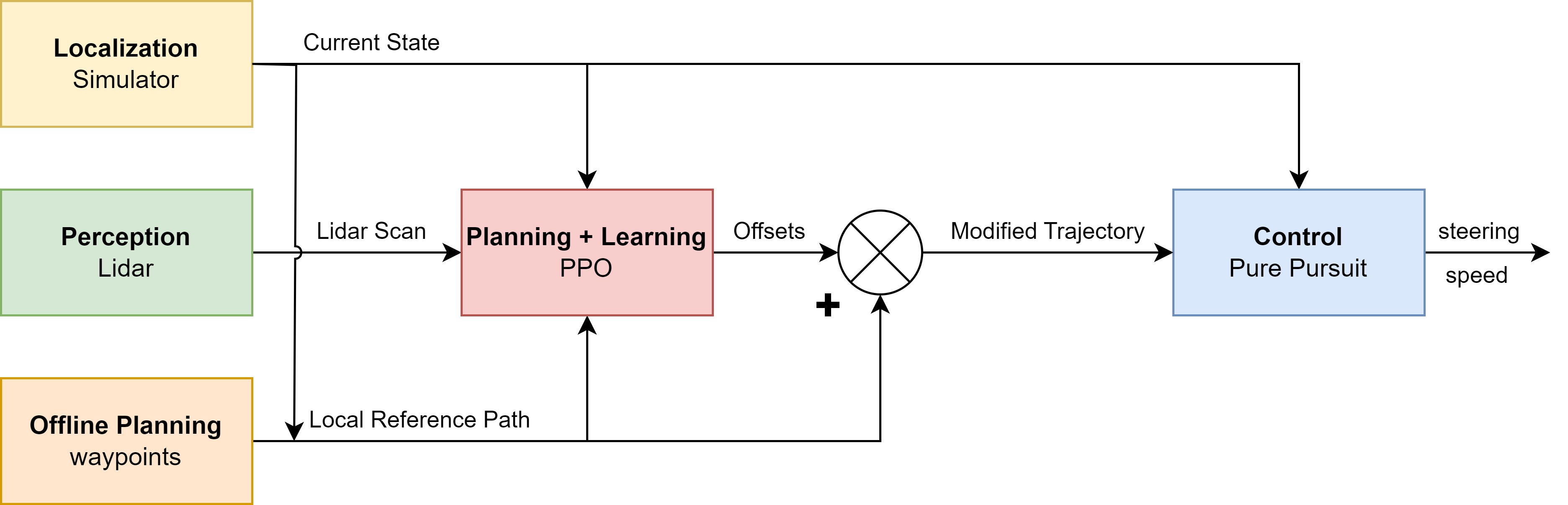} 
\caption{Structure of static obstacle nudging with path-planning-based PPO. Bootstrapped policy by BC is trained and tested using PPO to output lateral offsets for modifying paths, thereby avoiding obstacles. }
\label{fig:logic_RL_plot}
\end{figure}

Following the classic robotics workflow, we developed processes for path-tracking and static obstacle nudging, as depicted in Fig. \ref{fig:logic_IL_plot} and Fig. \ref{fig:logic_RL_plot}. 
The localization module provided the current state of the vehicle, which was then used as input for other modules, while the perception module produced lidar scans for sensing. 
The reference waypoints or raceline, was pre-generated as offline planning data to define the reference path. 
Behavioral Cloning (BC) and Proximal Policy Optimization (PPO) were employed within a "planning with learning" module, which processed three types of input data to generate lateral offsets for modifying the reference path. 
Finally, the controller utilized the modified path to compute steering and speed commands, allowing the vehicle to perform the desired maneuvers.

\subsection{Kinematic Bicycle Model}
Following the standard formalism of vehicle modeling, we simplified the Ackermann-steered vehicle as a single-track kinematic model \cite{automatic_steering}. 
In this model, the center of the rear axle represented the vehicle's position in 2D coordinates $(x, y)$, steering angle $\delta$, heading angle or orientation $\theta$, and wheelbase distance $L_{wb}$. Suppose no side slip, we denoted the car's longitudinal velocity as $v$. 
Therefore, we defined the vehicle state $\textbf{s}$ and desired action $\textbf{a}$ as follows:
\begin{equation*}
    \textbf{s} = [x, y, v, \theta], \quad \textbf{a} = [\delta_{des}, v_{des}].
\end{equation*}

\subsection{Lidar Scan}

The lidar scan data is defined as $\textbf{S}_p \in \mathbb{R}^k$, where $k$ denotes the number of lidar beams projected in the polar coordinates. To express the data more explicitly, we transformed $\textbf{S}_p$ into Cartesian frame of the vehicle:
\begin{equation*}
    \textbf{S} = T_p^c\textbf{S}_p. \tag{1}
\end{equation*}
$\textbf{S}$ is the lidar data in local Cartesian frame, and $T_p^c$ indicates the frame transformation.

\subsection{Waypoints}

Waypoint data is defined as $\textbf{w} \in (\mathbb{R}^5)^n$, where 
\begin{equation*}
    w_i = [x_i, y_i, v_i, \theta_i, \gamma_i], \ i = 1, ..., n
\end{equation*}
denotes the coordinates, the reference longitudinal speed, the heading angle, and the curvature of the $i$th waypoint in $n$ waypoints.
Following waypoints as a reference path, the vehicle proceeds tracking. 

In order to generate a global raceline for better reference, we applied an algorithm proposed in \cite{MINCURVE} as 
\begin{align}
\underset{[ \alpha_1 \cdots \alpha_n ]}{\text{min}} & \quad \sum_{i=1}^{n} \gamma_i^2(t) \tag{2} \\
\text{s.t.} & \quad \alpha_i \in [\alpha_{i,\text{min}}, \alpha_{i,\text{max}}]. \tag{3}
\end{align}
Through optimization parameter $\alpha_i$ that indicates the relative lateral position of track boundaries,
it minimizes the squared sum of every curvature $\gamma_i$ at time $t$ of spline between every adjacent waypoints. Hence, the optimal raceline is obtained as reference waypoints.

\subsection{Path Planning}

Suppose $H$ is the planning horizon and $\Delta t$ is the step time. Based on vehicle position $(x, y)$, we extracted the waypoint coordinates the car will go in $H \cdot \Delta t$ time with $H$ steps as $(x, y)_i, ..., (x, y)_j$. We interpolated these waypoints and evenly sample $H$ points including endpoints, and noted as horizon path $\textbf{t}_h$. Through homogeneous transformation $T^c_w$, local horizon path is obtained. After adding  offsets $\textbf{o}$ generated by learning methods, we transformed the path with offset back to the world frame through $T^c_w$ to get the modified path $\textbf{t}_m$ in world frame:
\begin{equation*}
\textbf{t}_m = T_c^w(T^c_w \textbf{t}_h + \textbf{o}). \tag{4} 
\end{equation*}

With the modified path $\textbf{t}_m$, the path-tracking controllers can take the modified path as new reference path to perform the obstacle nudging.

\subsection{BC for Path-Tracking}

BC can be formulated through Supervised Learning as eq. (5a), where the difference between the learned policy $\pi$ and expert demonstrations generated by the expert policy $\pi^*$ are minimized through loss function $\mathcal{L}$ with respect to some metric, and $\hat{\pi}^*$ is the approximated policy. 
\begin{align*}
\hat{\pi}^* 
    & = \underset{\pi}{\text{argmin}} \ \sum \mathcal{L}(\pi(\textbf{s}), \pi^*(\textbf{s}))
    \tag{5a} \\
    & = \underset{\pi}{\text{argmin}} \ \sum^t_{j=1}\sum^H_{i=1} |o_i|
    \tag{5b}
\end{align*}
Here, agent policy is $\pi(\textbf{s}) = \textbf{o}$, denotes the lateral offsets corresponding to $T^c_w \textbf{t}_h$.
Consider system dynamics and other factors, reference waypoints cannot be tracked perfectly even for the expert. However, the deviation is miscellaneous that can be ignored for simplification. 
L1 norm can be used as the metric to express the deviations, and the norm value in every time step can be added up as the loss function, shown as eq. (5b). 

By solving this optimization problem, the policy output is trained from penalizing random sampling and large deviations to converging to the expert performance with waypoints. This can be used for bootstrapping RL methods with rapid convergence and enhanced performance.

\subsection{PPO for Static Obstacle Nudging}

To achieve static obstacle nudging using waypoint data, lidar scans, and the current state with a bootstrapped model, we moved away from BC and other IL methods, because even expert demonstrations fall short for these tasks. Instead, we used PPO to train the policy through balancing exploration and exploitation. By focusing on policy performance without direct access to the environment, we chose policy optimization methods, specifically PPO, due to its proven performance and mature development.

In general, PPO optimizes the policy through
\begin{equation*}
\alpha_{k+1} = \text{arg} \underset{\alpha}{\text{max}} \ E \ [ \text{min} ( \frac{\pi_{\alpha}}{\pi_{\alpha_k}} A, \ g ) ], \tag{6} 
\end{equation*}
where $\alpha_k$ denotes the policy parameters during iteration $k$, $A$ is the advantage for the current policy $\pi_{\alpha_k}$, and $g$ is the clipping function.  
We referred to the implementation of CleanRL \cite{huang2022cleanrl} with further details. 

Optimizing the policy through PPO, instead of generating steering and speed commands like \cite{f110IL}, the policy outputs $\textbf{o}$, which deviate $\textbf{t}_h$ to be the modified path $\textbf{t}_m$. $\textbf{t}_m$ are then executed by Pure-Pursuit, a path-tracking method, thus achieving fixed obstacle nudging. 

\begin{figure*}[b]
    \centering
    \begin{subfigure}{0.235\textwidth}
        \centering
        \includegraphics[width=\textwidth]{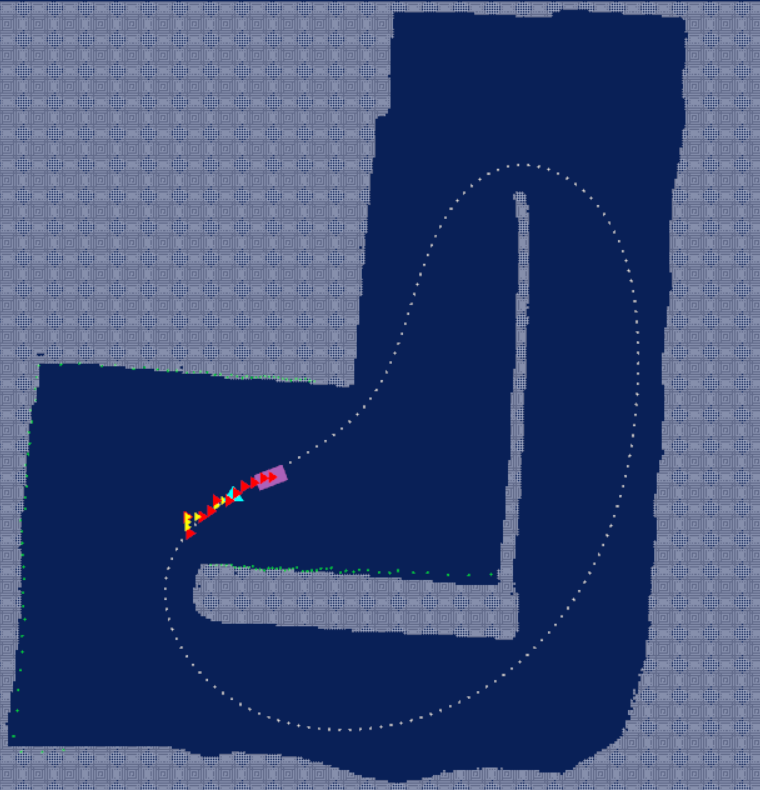}
        \caption{Pure Pursuit - 1M}
        \label{fig:shot_pp_1m}
    \end{subfigure}
    \hfill
    \begin{subfigure}{0.235\textwidth}
        \centering
        \includegraphics[width=\textwidth]{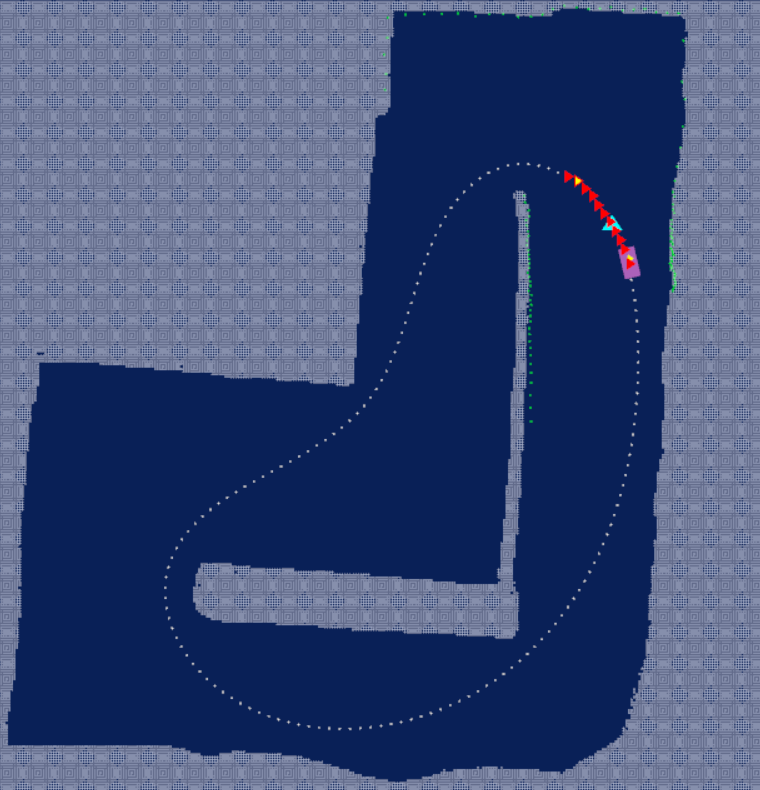}
        \caption{Pure Pursuit - 2M}
        \label{fig:shot_pp_2m}
    \end{subfigure}
    \hfill
    \begin{subfigure}{0.235\textwidth}
        \centering
        \includegraphics[width=\textwidth]{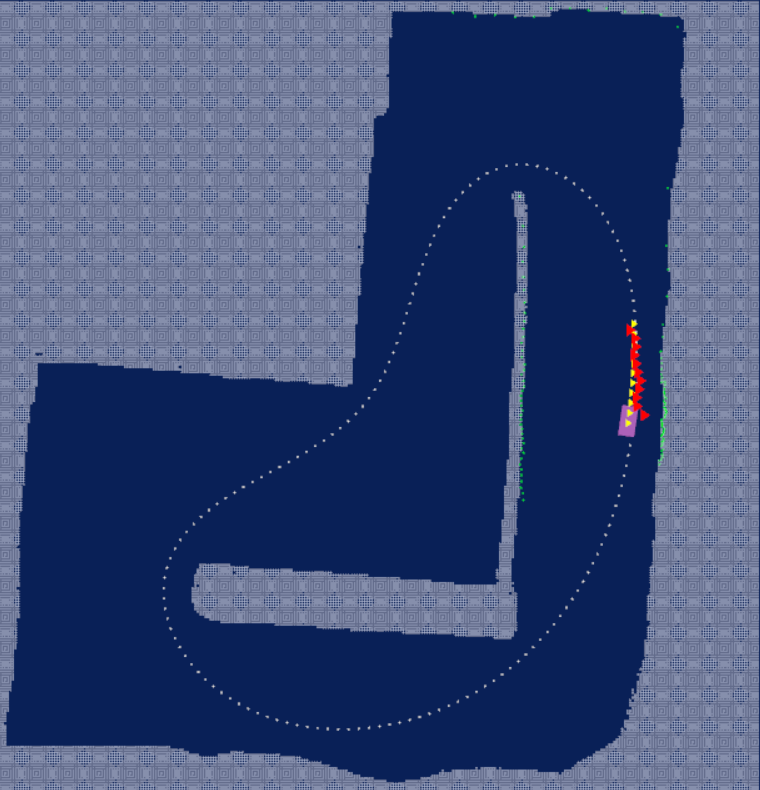}
        \caption{MPC - 1M}
        \label{fig:shot_mpc_1m}
    \end{subfigure}
    \hfill
    \begin{subfigure}{0.235\textwidth}
        \centering
        \includegraphics[width=\textwidth]{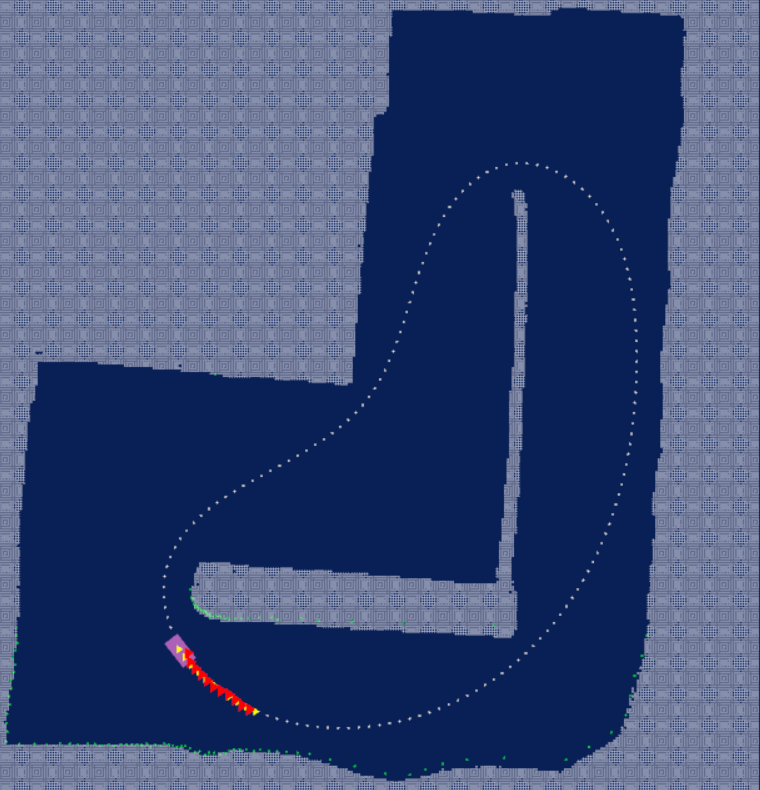}
        \caption{MPC - 2M}
        \label{fig:shot_mpc_2m}
    \end{subfigure}
    \caption{The BC performance for path-tracking with Pure Pursuit and MPC using different total timesteps. Modified paths and horizon paths are marked in red and yellow. Lookahead points are shown in cyan in Pure Pursuit plots. }
    \label{fig:bc_shots}
\end{figure*}

\subsection{Path-tracking Controllers}

\textbf{Pure-Pursuit} 
To pursuit the goal, a lookahead point is determined from a fixed lookahead distance $L$ towards the desired path.
Based on the kinematic bicycle model, the geometric relationship between $\delta$ and the turning radius $r$, and the arc curvature \(\gamma\) can be derived as:
\begin{equation*}
    \gamma = \frac{1}{r} = \frac{\text{tan}\delta}{L_{wb}} = \frac{2|e|}{L^2}, \tag{7}
\end{equation*}
where \(|e|\) is the cross-track error from the vehicle to the lookahead point, which coordinates are provided by the reference path. Therefore, $\delta$ can be solved to achieve tracking waypoints.  

\textbf{Model Predictive Control}
By building up a optimization problem with physical constraints and state dynamics, MPC can solve for a sequence of action. Define the state $z$ and the input $u$ based on the single-track kinematic model as
\begin{equation*}
    z = \begin{bmatrix}
        x & y & v & \theta \\
    \end{bmatrix}^T, \quad 
    u = \begin{bmatrix}
        a & \delta \\
    \end{bmatrix}^T, 
\end{equation*}
where $a$ is the desired vehicle acceleration. By discretization and linearization, system dynamics is derived, and objective along constraints are formulated as
\begin{align*}
\text{min} \quad
& \sum^{H-1}_{t=0} (z_t - z_t^r)^T Q_t (z_t - z_t^r) \\ 
& + (z_H - z_H^r)^T Q_H (z_H - z_H^r) \\
& + \sum^{H-1}_{t=0} (u_t - u_t^r)^T R_t (u_t - u_t^r)
  + \sum^{H}_{t=0} u_t^T R_d u_t \tag{8} \\
\text{s.t.} \quad 
& z_{t+1} = Ax_t + Bu_t + C \tag{9a} \\
& z_0 = z_{cur}, \ z_{min} \leq z_t \leq z_{max} \tag{9b, 9c} \\
& u_{min} \leq u_t \leq u_{max}, \ u'_{min} \leq u'_t \leq u'_{max}. \tag{9d, 9e} 
\end{align*}
In (8), $Q_t, \ Q_H$ stand for step and final penalty matrix of $z$, $R_t, \ R_d$ are step and differential penalty matrix of $u$. (9a) indicates the system dynamics with system matrices $A, \ B, \ C$ \cite{raj2012}. (9b) requires the initial condition is the current state $z_{cur}$, (9c, 9d) limit $z_t, \ u_t$ respectively, and (9e) constrains the difference of $u_t$. The vehicle executes solved $u_t$, which is $a$ and $\delta$ for tracking.

\section{EXPERIMENTS}

In this section, we present detailed implementation and verification through experiments in simulation scenarios. We demonstrate that the path-planning-based BC and PPO bootstrapped by BC achieve great performance in path-tracking and static obstacle nudging respectively. The implementation of code, video links, instructions, and additional resources are available at \url{https://github.com/derekhanbaliq/Planning-with-Learning}.

\subsection{Experimental Setup} 

We conducted development and verification on f1tenth\_gym, a simulation environment of F1TENTH based on Gym. f1tenth\_gym offers a robust closed-loop simulation framework facilitating rapid implementation. To incorporate learning-based methods, we integrated the single-file PPO implementation from CleanRL \cite{huang2022cleanrl} into f1tenth\_gym, enabling the use of PPO.

The experimental setup was deployed following various considerations. 
Hokuyo lidar scan data is downsampled from 1080 to 108 beams, as described in \cite{f110IL}, to reduce the model's input dimension. 
The path prediction time is set to 1 second for path-tracking and 2 seconds for bootstrapping to static obstacle nudging. 
The planning horizon was defined as $H = 10$. 
The vehicle's initial pose was randomized along the reference waypoints while maintaining the same orientation and avoiding collisions. 
For Pure Pursuit, the vehicle was configured with a fixed lookahead distance of $L = 0.8 \ \text{m}$ and a constant speed of $2 \ \text{m}/\text{s}$ to ensure stable performance. 
Correspondingly, the speed of the MPC was capped at a maximum of $2 \ \text{m}/\text{s}$. 
Additionally, the control frequency was set to $10 \ \text{Hz}$. 

The agent model was actor-critic, compatible with the PPO design. 
The critic network learned the value function, which estimates the expected reward of being in a particular state; while the actor network outputted the mean of the action distribution. 
Both networks shared the same structure, consisting of a $4 \times 256$ multi-layer perceptron. 
The learning rate was set to $3 \times 10^{-4}$, the generalized advantage estimation was set to $0.95$, and the discount factor was set to $0.99$. The maximum gradient norm was established at $0.5$ to prevent gradient explosions. 
To design the reward function, we calculated the reward $r$ based on the following values: the longevity of the vehicle, which was accumulated through the number of steps $n$ ($0.01 \ \text{s}$ each) without crashing or finishing laps, the 2-norm penalty of offsets $\textbf{o}$, and a collision penalty $C = 1000$. The reward function was formulated as follows: 
\begin{equation*}
    r = 100 \cdot n - ||\mathbf{o}||_2 - C. \tag{10}
\end{equation*}

\begin{figure}[t!]
\centering
\includegraphics[width=\columnwidth]{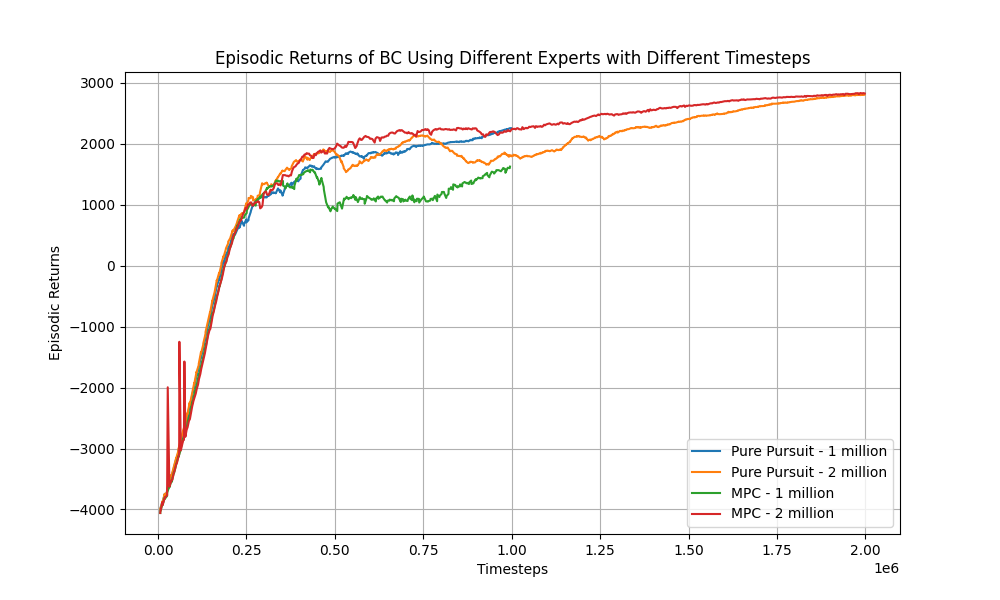} 
\caption{Episodic returns of BC for path-tracking with Pure Pursuit and MPC for different training timesteps. }
\label{fig:epi_returns}
\end{figure}

\subsection{Path-Tracking with BC}

We trained agents using demonstration data that utilized Pure Pursuit and MPC controllers separately. 
To compare the performance of the two experiments, we set the prediction time to 1 second, consistent with the MPC setup, consistent with the MPC setup. 
The total number of training time steps for BC was set to 1 million, while 2 million steps were used for comparative analysis. 

For the training result, the episodic returns, shown in Fig. \ref{fig:epi_returns}, illustrate the rapid convergence and learning efficiency of BC. 
With additional training, the models have higher return values than the models trained with 1 million. Besides, the 2 million models reach to nearly the same return value, indicating the method is compatible with different path-tracking controllers.
Fig. \ref{fig:bc_shots} illustrates the modified paths are close to reference waypoints, showing the great performance of path-tracking through BC. 
With extended training, the paths predicted by agents become smoother with less deviation, which is also evident depicted in Fig. \ref{fig:epi_returns}.

\begin{figure}[b]
    \centering
    \begin{subfigure}{0.235\textwidth}
        \centering
        \includegraphics[width=\textwidth]{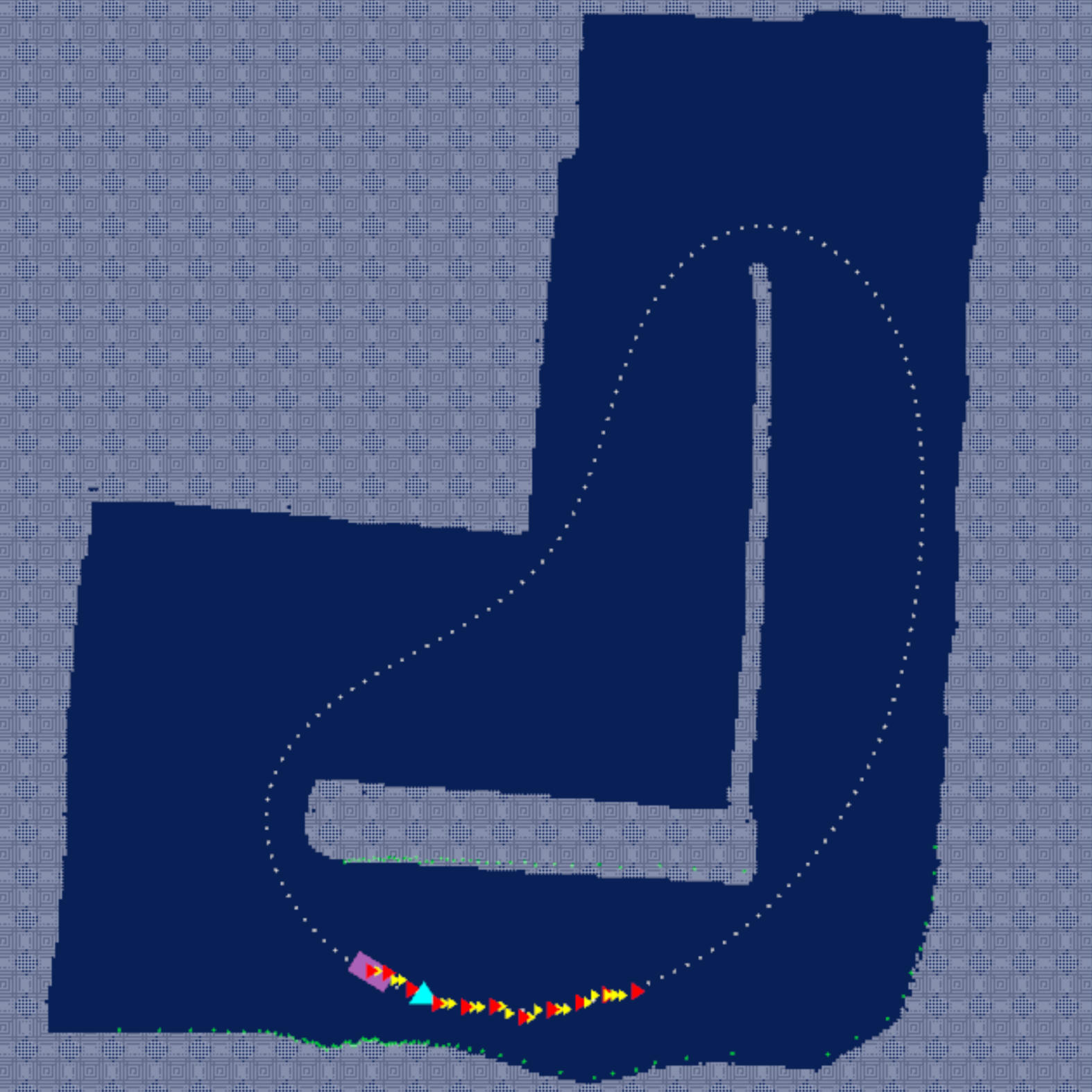}
        \caption{2s Prediction - 1M}
        \label{fig:1s_bt_pp}
    \end{subfigure}
    \hfill
    \begin{subfigure}{0.235\textwidth}
        \centering
        \includegraphics[width=\textwidth]{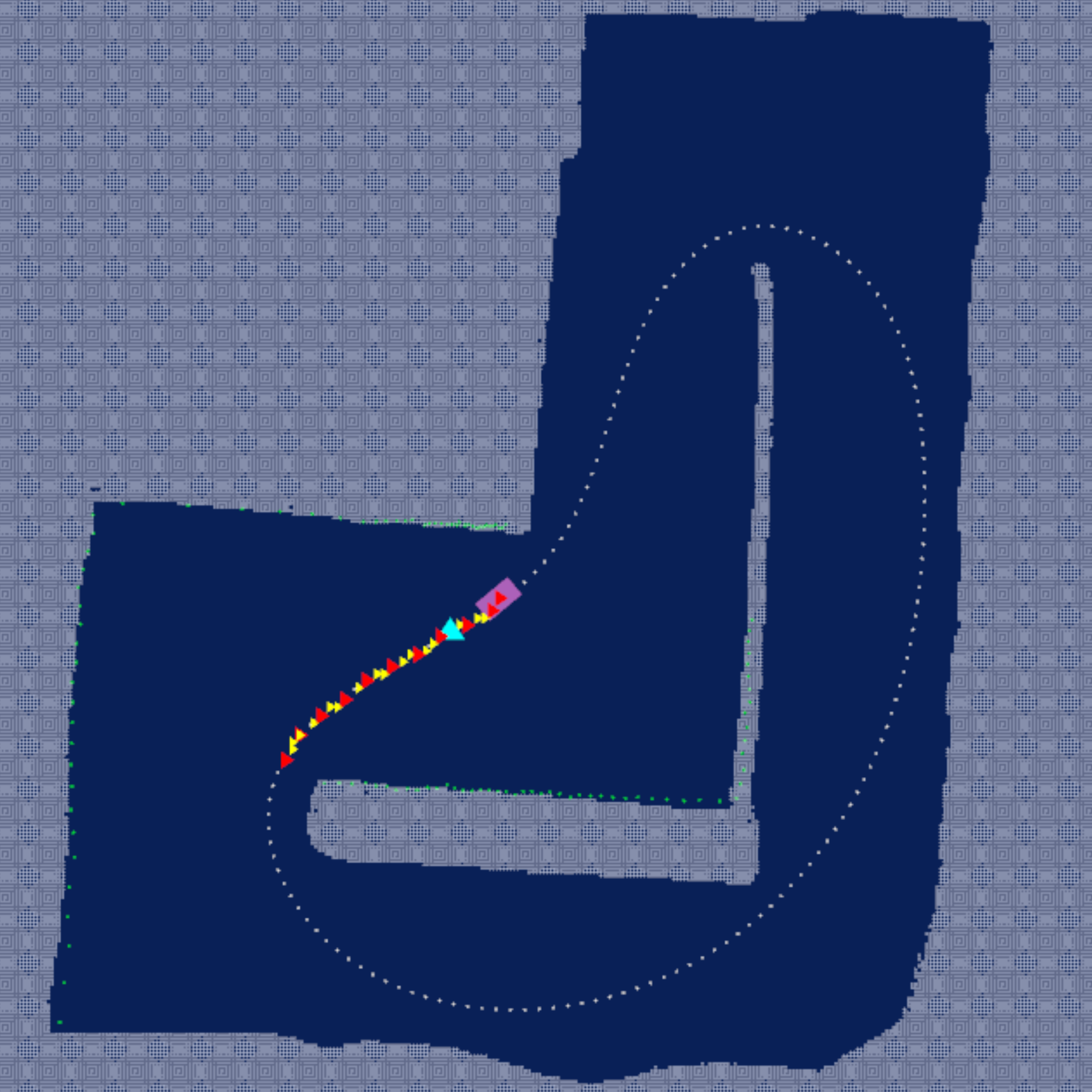}
        \caption{2s Prediction - 2M}
        \label{fig:2s_bt_pp}
    \end{subfigure}
    \caption{The BC performance for bootstrapping static obstacle nudging with PPO in the obstacle-free map.}
    \label{fig:nudging_bt_pp}
\end{figure}

\subsection{Static Obstacle Nudging with PPO}

Static obstacle nudging was achieved by adjusting selected reference waypoints through lateral offsets $\textbf{o}$. 
To illustrate path deviations from the horizon path, we bootstrapped models using BC with Pure Pursuit demonstrations. The models are configured with a $2 \ \text{s}$ prediction time and trained over a total of 1 and 2 million timesteps respectively. 
The training results are depicted as Fig. \ref{fig:nudging_bt_pp}, which shows that both two models provide stable performance. We used 1M model for the following nudging experiments. 
To create an obstacle map, we modified the original obstacle-free map by adding 2, 3, and 4 static obstacles respectively. Each obstacle box measures $7 \times 7$ pixels, approximately  $35 \times 35 \ \text{cm}$, which side length is a full width of an F1TENTH car in real scenarios. The obstacles are strategically placed along the waypoints to evaluate the vehicle's nudging performance. 

\begin{figure*}[t]
    \centering
    \begin{subfigure}[b]{0.32\textwidth}
        \centering
        \includegraphics[width=\textwidth]{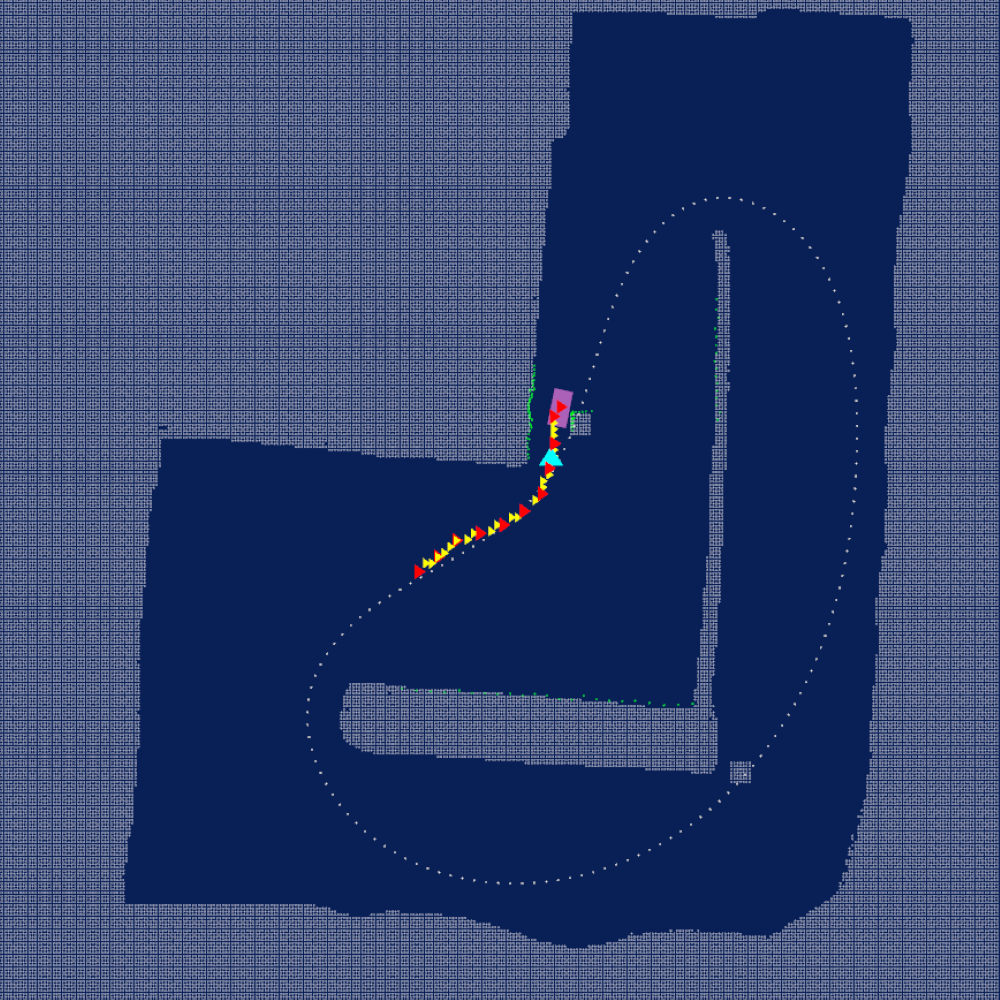}
        \label{fig:shot_2_obs}
    \end{subfigure}
    \hfill
    \begin{subfigure}[b]{0.32\textwidth}
        \centering
        \includegraphics[width=\textwidth]{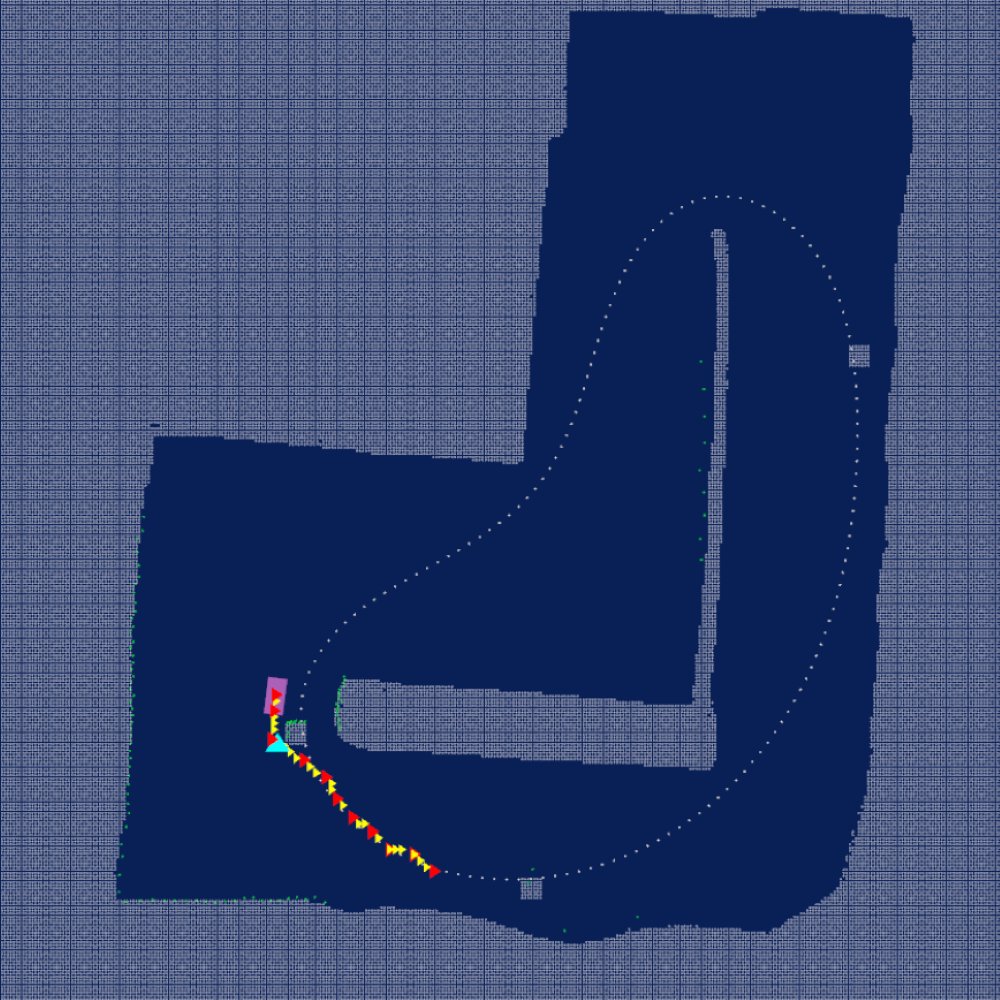}
        \label{fig:shot_3_obs}
    \end{subfigure}
    \hfill
    \begin{subfigure}[b]{0.32\textwidth}
        \centering
        \includegraphics[width=\textwidth]{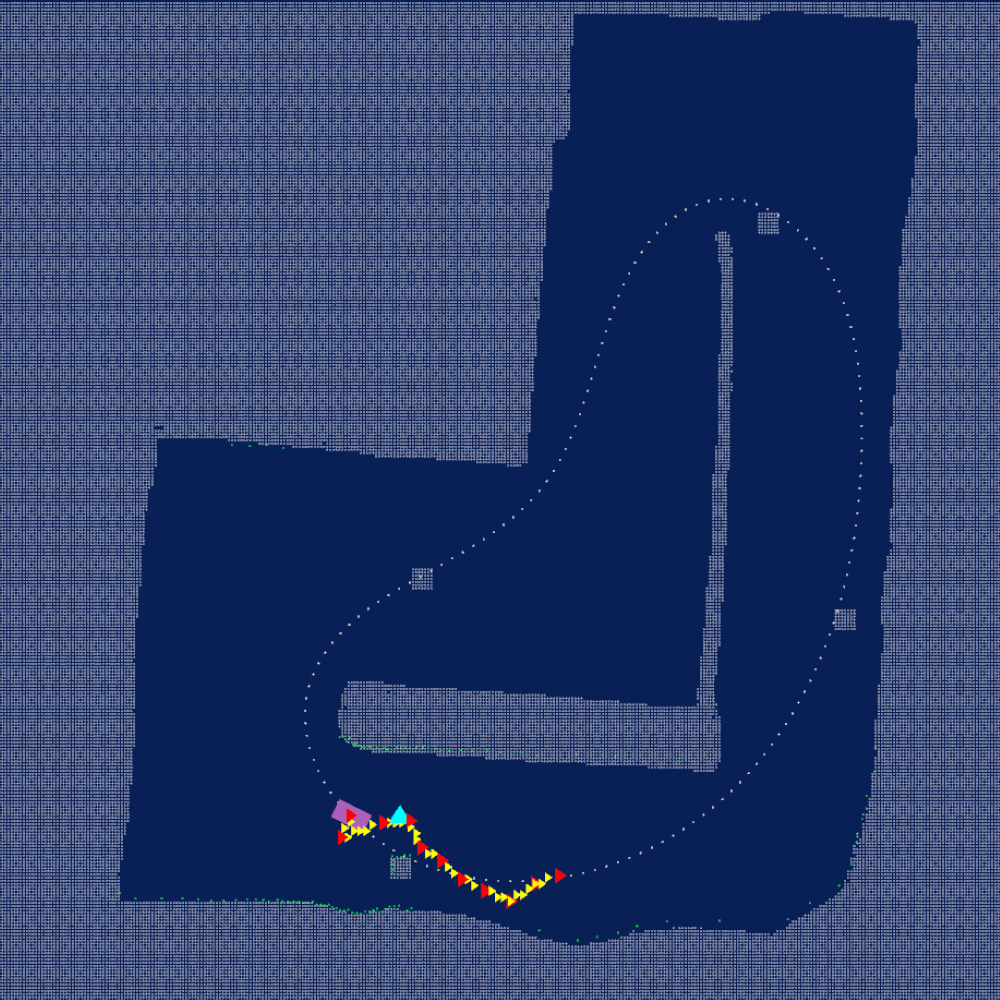}
        \label{fig:shot_4_obs}
    \end{subfigure} \\
    \begin{subfigure}{.32\textwidth}
        \centering
        \includegraphics[width=\textwidth]{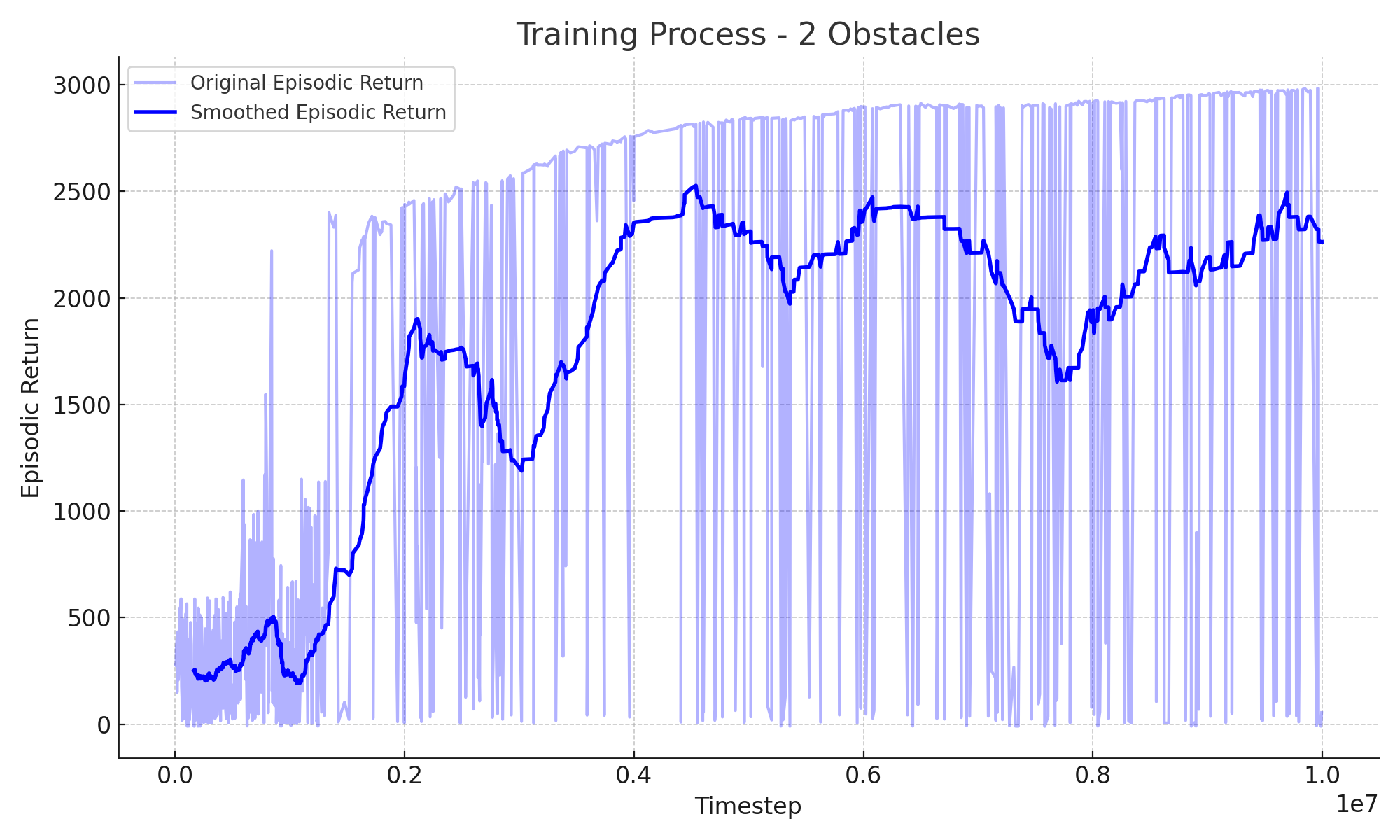}
        \caption{Nudging - 2 Static Obstacles}
        \label{fig:return_2_obs}
    \end{subfigure}
    \hfill
    \begin{subfigure}{.32\textwidth}
        \centering
        \includegraphics[width=\textwidth]{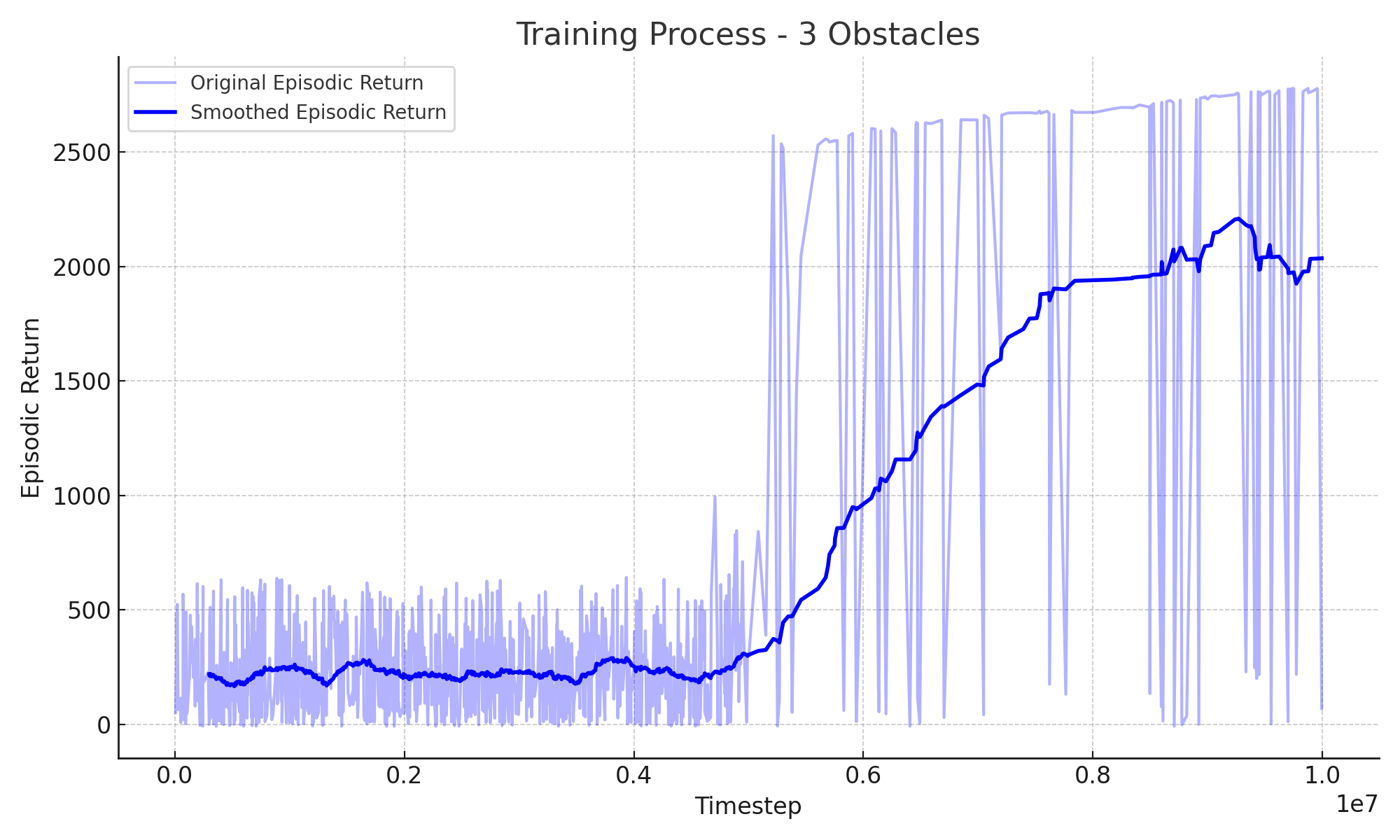}
        \caption{Nudging - 3 Fixed Obstacles}
        \label{fig:return_3_obs}
    \end{subfigure}
    \hfill
    \begin{subfigure}{.32\textwidth}
        \centering
        \includegraphics[width=\textwidth]{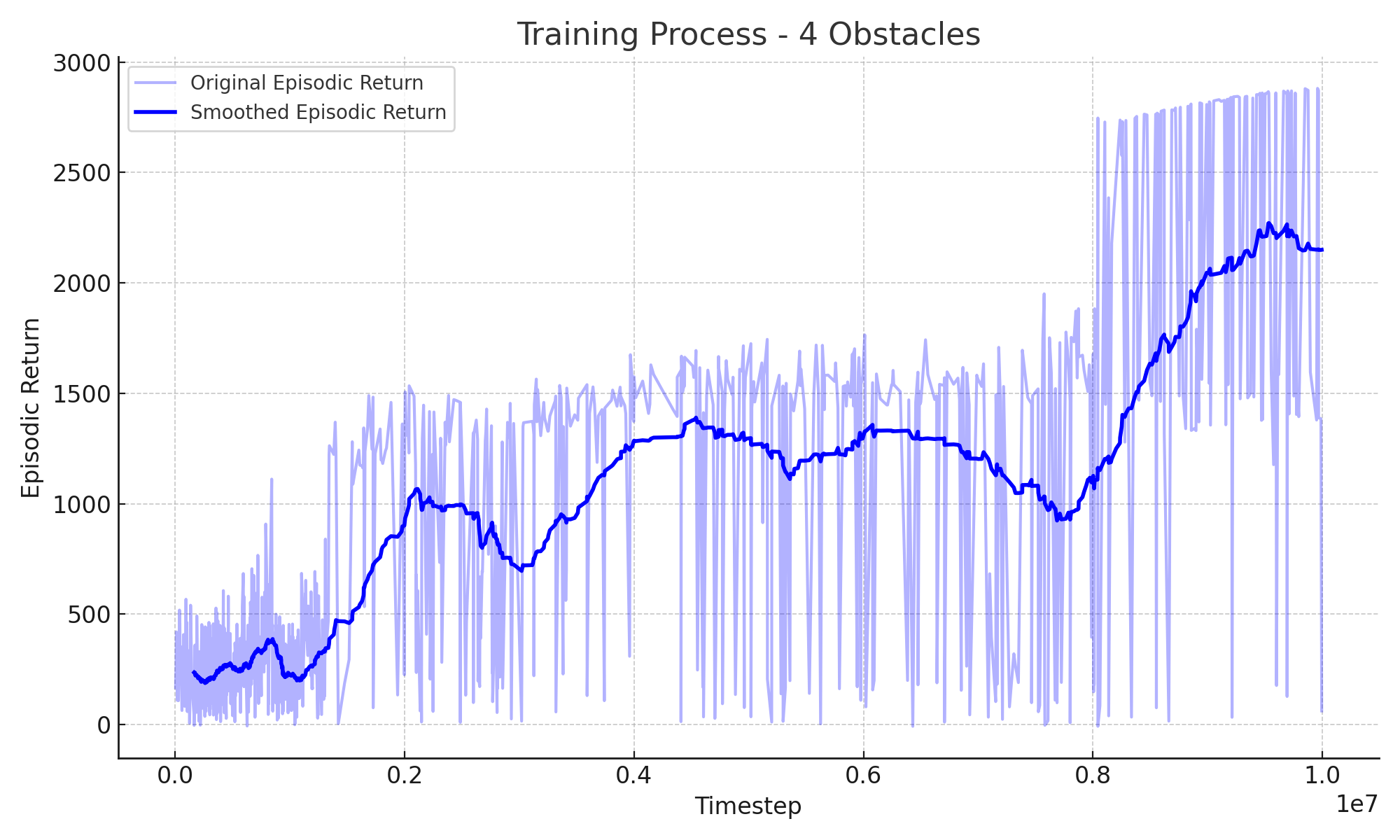}
        \caption{Nudging - 4 Fixed Obstacles}
        \label{fig:return_4_obs}
    \end{subfigure}
    \caption{Static obstacle nudging and corresponding episodic returns using BC-bootstrapped PPO with Pure Pursuit validated with different fixed obstacle setups. Modified paths are marked in red triangles and yellow lines, and cyan triangles denote the lookahead points. }
    \label{fig:obs_avoid_shots}
\end{figure*}

We used a total of 10 million timesteps to train the aforementioned model. 
The reward function was also modified with extra $||\mathbf{o}||_1$ term for extra penalty. 
The actual performances are depicted in Fig. \ref{fig:obs_avoid_shots}. 
The lateral offsets $\textbf{o}$ successfully modified the horizon path $\mathbf{t}_h$ which was truncated from reference waypoints to obtain a modified path $\textbf{t}_m$. This modified path was then employed by Pure Pursuit, the path-tracking controller, to execute actual maneuvers. In this context, Pure Pursuit calculates a lookahead point to guide the vehicle in avoiding obstacles using $\textbf{t}_m$. 
The corresponding episodic returns shown in Fig. \ref{fig:obs_avoid_shots} increased throughout the training process, illustrating the convergence of the policies. 

\section{Conclusion}

In this work, we introduced a method that uses BC and PPO algorithms for path planning tasks, specifically path-tracking and static obstacle nudging. 
The experiments in the F1TENTH Gym environment validated that this method effectively performs both path-tracking using BC and fixed obstacle nudging using PPO bootstrapped by BC. 
The development demonstrated the efficacy of integrating learning into planning and highlights the practical benefits of combining RL with IL in the field of autonomous driving.
Future could focus on reducing the sim-to-real gap for robust deployment, stronger generalizability to various kinds of dynamic obstacles, and decoupling the planning from decision-making to physics-constrained motion planning. 

\section{Acknowledgement}

The authors express their sincere gratitude to Yi Shen from the Robotics Department at the University of Michigan for his helpful discussions and advice.



\bibliography{main.bib}

\end{document}